%% file: main.tex
\documentclass[12pt]{article}

\usepackage{algpseudocode}
\usepackage{algorithm}
\usepackage{amsmath,amsthm}
\usepackage{color}
\usepackage{graphicx}
\usepackage{amssymb}
\usepackage{cleveref}
\usepackage{csquotes}
\usepackage[T1]{fontenc}
\DeclareMathOperator*{\argmax}{arg\,max}
\DeclareMathOperator*{\argmin}{arg\,min}

\title{Vulnerability Analysis of Transformer-based Optical Character Recognition to Adversarial Attacks}

\author{Lucas Beerens\thanks{School of Mathematics and The Maxwell Institute for Mathematical Sciences, University of Edinburgh, EH8 9BT, UK} \and
Desmond J. Higham\thanks{School of Mathematics and The Maxwell Institute for Mathematical Sciences, University of Edinburgh, EH8 9BT, UK}}

\begin{document}
\maketitle
\input{sec/0_abstract}    
\input{sec/1_intro}

\input{sec/2_related_work}
\input{sec/3_method}
\input{sec/4_experiment}
\input{sec/5_conclusion}

\section*{Funding}
LB was 
  supported by the MAC-MIGS Centre for Doctoral Training under EPSRC grant EP/S023291/1.
  DJH was supported 
  by EPSRC grants EP/P020720/1 and EP/V046527/1.

\section*{Data Statement}
Code for the experiments presented here will be made available upon 
publication.

\bibliographystyle{siam}
\bibliography{main}

\end{document}

%% file: sec/0_abstract.tex
\begin{abstract}
Recent advancements in Optical Character Recognition (OCR) have been driven by transformer-based models. OCR systems are critical in numerous high-stakes domains, yet their vulnerability to adversarial attack remains largely uncharted territory, raising concerns about security and compliance with emerging AI regulations.
In this work we present a novel framework to assess the resilience of Transformer-based OCR (TrOCR) models. 
We develop and assess algorithms for both targeted and untargeted attacks. For the untargeted case, we measure the Character Error Rate (CER), while for the targeted case we use the success ratio. We find that TrOCR is highly vulnerable to untargeted attacks and somewhat less vulnerable to targeted attacks. 
On a benchmark handwriting data set, untargeted attacks can cause a CER of more than 1 without being noticeable to the eye.  With a similar perturbation size, targeted attacks can lead to success rates of around $25\%$---here we attacked single tokens, requiring TrOCR to output the tenth most likely token from a large vocabulary.
\end{abstract}

%% file: sec/1_intro.tex
\section{Introduction}
\label{sec:intro}

Optical Character Recognition (OCR)  is the conversion of images of printed or handwritten text into machine-interpretable text. This has many applications, ranging from document digitization and automated data entry to facilitating accessibility for visually impaired individuals. OCR has gone through several decades of development \cite{memon2020handwritten}. In the last decade, the resurgence of neural networks has led to significant improvements in OCR capabilities. Both text detection and text recognition were able to take advantage of the advances in 
convolutional neural networks (CNNs). In particular, text recognition models have been formulated as encoder-decoder systems, where the encoder uses CNNs, and the decoder uses
recurrent neural networks
(RNNs) \cite{shi2016end}.

\begin{figure}
    \centering
    \includegraphics[width=\linewidth]{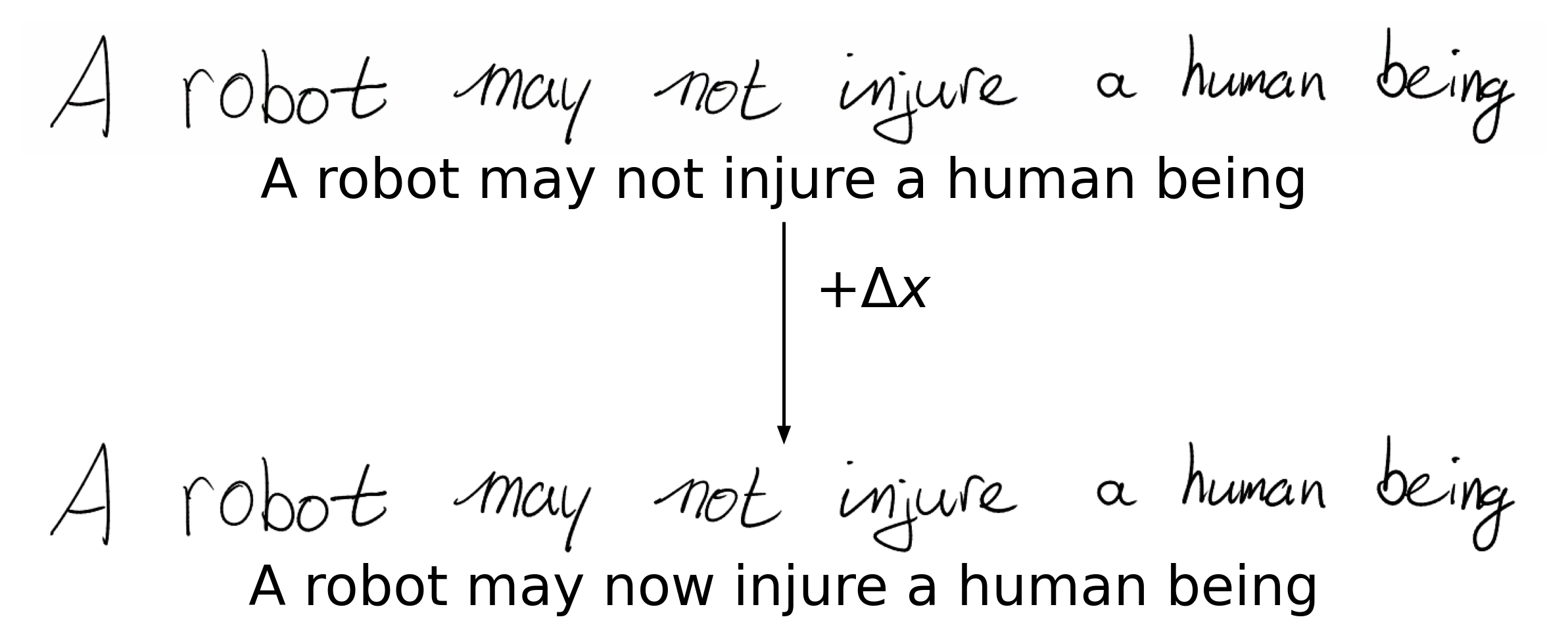}
    \caption{Adversarial attack created by tailoring a Carlini \& Wagner attack \cite{carlini2017towards} to the TrOCR setting. 
    The full algorithm is described in \cref{Sec:Method/CW}. 
    Here a targeted version of the attack is used---we aim for the optical character recognition system to create a specific output on the perturbed sentence. The original 
    sentence (upper), handwritten by the first author and scanned, was correctly recognized by TrOCR as `A robot may not injure a human being'. After the 
    imperceptible adversarial perturbation is added (lower), TrOCR incorrectly outputs `A robot may now injure a human being'. This was the output target supplied to the attack algorithm. The original sentence is inspired by the first law of robotics by Isaac Asimov \cite{asimov1950robot}.}
    \label{fig:robotcomp}
\end{figure}

In recent years, the widespread adoption of transformers \cite{attentionPaper}, originally designed for natural language processing (NLP), has yielded remarkable advances in various domains, including NLP \cite{kalyan2021ammus, zhao2023survey, acheampong2021transformer}, Computer Vision (CV) \cite{khan2022transformers, lahoud20223d}, and speech processing \cite{latif2023transformers}. Utilization of transformers in NLP and CV has now been extended to OCR. First, this was done while keeping CNNs in the backbone \cite{diaz2021rethinking}. Later, a model without CNN backbone was created called Transformer-based Optical Character Recognition (TrOCR) \cite{li2023trocr}.

However, amidst these achievements, a critical concern remains. Many AI systems can be manipulated by mischievous, malicious or criminal third parties.  
Adversarial attacks, which, for example, introduce imperceptible perturbations into clean images to deceive deep learning 
classification models, have drawn substantial attention in the past decade \cite{szegedy2013intriguing, harness, liang2022adversarial}. 
A cat-and-mouse game between attack and defence strategies is continuing, with Carlini 
\cite{Carlini2023llm} observing that 
\begin{displayquote}
``Historically, the vast majority of adversarial defenses published at top-tier conferences \ldots are quickly broken.''
\end{displayquote}
Given that recent TrOCR technology offers benefits in many  
high-risk and safety-critical areas, including legal, financial and healthcare applications, it is crucial to understand 
its vulnerability to adversarial attack. This issue, which currently remains unaddressed, motivates our work.

At a policy-making level, 
concerns about AI in general have led to a call for regulation to ensure the safety and transparency of AI. An example of such regulation is the 
proposed EU AI act, which was first released in April 2021 \cite{euaiact} and amended in June 2023 \cite{euaiactamendment}. Article 15 – paragraph 4 – subparagraph 1 of the amended proposal states: 
\begin{displayquote}
    ``High-risk AI systems shall be resilient as regards to attempts by unauthorised third parties to alter their use, behaviour, outputs or performance by exploiting the system vulnerabilities."
\end{displayquote}
Highlighting one of many application domains, we note that 
Annex III – paragraph 1 – point 3 – point b of this same proposal mentions that the following class of AI systems is deemed high risk:
\begin{displayquote}
    ``AI systems intended to be used for the purpose of assessing students in educational and vocational training institutions and for assessing participants in tests commonly required for admission to those institutions." 
\end{displayquote}
OCR systems can be used as part of an AI pipeline designed for student assessment \cite{rajesh2019digitized}. Therefore, in this and other high-risk domains,  understanding 
and addressing their weaknesses is imperative before deployment. 

Our main contributions in this work are as follows. 
\begin{itemize}
    \item We create a novel framework to assess the resilience of Transformer-based Optical Character Recognition.
    \item We tailor strategies to TrOCR by building on established image classification attacks, resulting in attacks such as the one in \cref{fig:robotcomp}.
    \item We assess the vulnerability of TrOCR to both targeted and untargeted attacks on the IAM handwriting dataset by comparing multiple attack algorithms.
\end{itemize}

%% file: sec/2_related_work.tex
\section{Related Work}
\label{sec:related_work}
\subsection{Adversarial Attacks on CNNs}
The first adversarial attacks were carried out on fully connected neural networks and CNNs in the context of image classification \cite{szegedy2013intriguing, harness}. Subsequently, these ideas have been extended to other areas, including NLP, and other architectures within CV 
 \cite{wang2022nlp,zhang_2022_CVPR,wei_2022_CVPR}.

Adversarial attacks can be classified as white-box or black-box. The first adversarial attacks fell into the white-box category: they had access to
information about the neural network, including gradients with respect to input components. Examples of widely-used white-box methods include L-BFGS \cite{szegedy2013intriguing}, FGSM \cite{harness}, and DeepFool \cite{moosavi2016deepfool}. 
These techniques are readily generalized to black-box attacks, where less information about the targeted neural network is available \cite{papernot2017practical}. A common approach in black-box attacks is to use gradient-based white-box attack algorithms optimized for transferability, replacing exact gradients with approximations. Two such types of attack are transfer-based and query-based. In the transfer-based case, the gradient of a surrogate model is used \cite{papernot2017practical}. In the query-based case, the gradient is estimated by querying the model with a number of inputs \cite{chen2017zoo, bai2023query}.
Hence, white-box methods are relevant to the black-box context. 

Many strategies have been employed in the development of white-box techniques. These can be broadly categorized as gradient optimization, constrained optimization, and generative models. Within these categories, a targeted attack aims for a specific, new output class, 
whereas a non-targeted attack aims for any change of output class. 

Since attacks typically aim for the largest output change for a given (small) 
size of input change, the norm used to measure perturbations is a key ingredient. 
Most attacks use $\ell_0$, $\ell_2$, or $\ell_\infty$ norms \cite{sharif2018suitability}. 
When the objective is to hide a perturbation in the image, these norms can be interpreted as proxies for the visibility of the attack. There are also 
other forms of disguise, such as attacking speech recognition systems using psychoacoustic hiding \cite{schonherr2018adversarial} or attacking handwritten digit classification by hiding the perturbation in the ink \cite{beerens2023adversarial}. 
In this work, we will be using the $\ell_2$ norm to measure perturbation size.

\subsection{Optical Character Recognition}
Attacks on CNN models have focused mainly on classification. Here, we consider the TrOCR model, which is designed for Optical Character Recognition. 
Two types of input can be considered: handwritten and printed text. Printed text is 
the simpler case due to the uniformity of dimensions, positions, and fonts. We focus on handwritten text. Handwritten OCR may also 
be divided into two cases: online and offline. In the former case, the system can use all information that can be captured while the text is written. This means that the system knows the direction and order of the pen strokes while not having any issues with stroke thickness. In the offline case, this information is not available and only the resulting image can be used. We focus on the more widely applicable offline case. 

Typically, OCR refers to the entire pipeline from data acquisition to post-processing. This pipeline can be separated into the following steps: acquisition, pre-processing, segmentation, feature extraction, classification, and post-processing. The first three steps of the text processing pipeline can be grouped together as text detection, while the last three can be grouped together as text recognition. We focus on 
segmented examples from the IAM Handwriting Database \cite{marti2002iam}. (Figure~\ref{fig:robotcomp} uses a scan of a sentence that was handwritten by the 
first author.)



\subsection{Transformer-based OCR}
TrOCR is an end-to-end text recognition system, which takes textline segmented images and performs text recognition \cite{li2023trocr}. It does not use any convolutional layers in its architecture. Instead, it uses the transformer architecture \cite{attentionPaper}. TrOCR consists of three steps. First the image is resized to a fixed resolution and split into a sequence of image patches. These are used as input to a pre-trained Vision Transformer model, which encodes this sequence of patches \cite{dosovitskiy2020image}. This is decoded by a pre-trained transformer language model, which generates wordpiece tokens based on the image and the context generated before. TrOCR was pre-trained with synthetic data and fine-tuned with human-labeled datasets. Different combinations of encoder and decoder architectures are used to create versions of different sizes. 
$\text{TrOCR}_\text{SMALL}$ uses $\text{DeiT}_\text{SMALL}$ for the decoder \cite{touvron2021training} and MiniLM for the encoder \cite{wang2020minilm}, giving a total of $62$M parameters, compared with 
 $334$M and $558$M parameters, respectively,
 for
$\text{TrOCR}_\text{BASE}$ and $\text{TrOCR}_\text{LARGE}$.
We focus on $\text{TrOCR}_\text{SMALL}$ which allows us to conduct more extensive experiments; this choice is supported by 
empirical \cite{papernot2016distill, xiao2019train,rodriguez2022bio}
and theoretical \cite{thwg21} results showing that smaller models are more resilient to attack, in general.

\subsection{Adversarial attacks on transformer models}
The rise of transformers has led to investigations into their vulnerability to adversarial attacks. In particular, Visual Transformers (ViTs) have been examined, with results suggesting that they are generally more robust than CNNs \cite{heo2023exploring, shao2021adversarial, aldahdooh2021reveal}. However, the attacks originally designed for CNN systems remain somewhat effective against ViTs and some recent work has focused on improving transferability \cite{wei2022towards, mahmood2021robustness, zhang2023transferable}.

Transformers have also been used in NLP to build language models such as BERT \cite{devlin2018bert} and MiniLM \cite{wang2020minilm}. Such models can be used to create systems for a variety of applications, such as machine translation, sentiment analysis, and summarization. 
Models that generate tokens are the most relevant to our work, as they are broadly similar to the TrOCR decoder. Adversarial attacks have been found that  generate effective adversarial examples for these types of models \cite{morris2020textattack}. 

Since TrOCR has image inputs, like the ViTs, but has generative text output, like language transformer models, it is substantially different from both cases.  Adversarial attacks on TrOCR have been investigated in the context of adding diacritics to printed text \cite{boucher2023vision}. This is markedly different from our work since it specifically uses printed text, and the perturbations are within the ASCII character set. To the best of our knowledge, adversarial attacks on models such as TrOCR have not been investigated before in the context of small perturbation to input images.

%% file: sec/3_method.tex
\section{Method}
In \cref{Sec:Method/trocr} we present a framework for dealing with TrOCR mathematically. Then we introduce adversarial attacks against TrOCR based on various existing methods in \cref{Sec:Method/FGSM,Sec:Method/DeepFool,Sec:Method/BE,Sec:Method/CW}. Finally, we discuss the implementation in \cref{Sec:Method/Implementation}. \cref{Sec:Experiments} presents experimental results and \cref{Sec:Conclusion} gives conclusions. 

\subsection{TrOCR}\label{Sec:Method/trocr}
In creating adversarial attacks against TrOCR, we will be using gradients of functions applied to the TrOCR output. Therefore, it is important to describe this output carefully. Images are represented by a real-valued vector $x$. In the inference process, images are first passed through a preprocessor to give them a similar shape and distribution. Then, they are passed through an encoder that encodes image patches into vectors. Let us call the output of these first two steps $E(x)$. The decoder will iteratively generate tokens from this. It starts with a special beginning-of-sentence (bos) token. In every iteration, the output is the current sequence of tokens shifted to the left (so without the bos token), with the new token added at the end. Once the end-of-sequence (eos) token has been generated or the maximum amount of tokens has been reached, the process stops. The output is a sequence of logit vectors. The elements in the vectors correspond to  vocabulary tokens. Applying the softmax function would result in probabilities, but we do not need to consider such a step. The element with the maximum value determines the token assigned. The inputs and outputs have the form
\begin{equation}
    (E(x),\left((\text{bos}),t_1,\hdots,t_k)\right) \mapsto (\mathbf{t}_1,\hdots,\mathbf{t}_{k+1}),
\end{equation}
where for all $i$ in $\{1,\hdots k+1\}$ we have
\begin{equation}
    \mathbf{t}_i = \begin{pmatrix}
        l_{i,1}\\
        \vdots\\
        l_{i,v}
    \end{pmatrix},
\end{equation}
and $t_i=\argmax_j\{l_{i,j}\}$. Here, $v$ is the size of the vocabulary. Vector $\mathbf{t}_i$ does not change in subsequent iterations. This means that the logits for the generation of all tokens can be obtained by taking the final output. The final logit matrix is denoted by $F(x)$ with entries $F^i_j(x)$, where $i$ is the position of the token and $j$ is the token class in the vocabulary.

\subsection{FGSM}\label{Sec:Method/FGSM}
The Fast Gradient Sign Method (FGSM) \cite{harness} is an early and widely used adversarial attack strategy. Let us first consider the untargeted version of FGSM. The gradient of the loss function with respect to the correct label points in the direction where it will increase the most, locally. FGSM looks at the sign of this gradient for every pixel and takes a step in that direction. Increasing the loss means moving away from the correct answer. 
The attack is computed as 
\begin{equation}
    \Delta x = \epsilon  \operatorname{sign}\left(\nabla_x \mathcal{L}(x,L_\text{out})\right),
    \label{eq:Dx}
\end{equation}
where $\mathcal{L}$ denotes the loss function used in training with respect to the output labels $L_\text{out} = (t_0,\hdots,t_k)$ and $\epsilon$ denotes some allowed perturbation size. For small $\epsilon$, the perturbation $ \Delta x$ in (\ref{eq:Dx}) causes the largest change under the constraint 
$\| \Delta x \|_{\infty} \le \epsilon$.

FGSM can be used in a targeted manner by instead computing the loss with respect to some target labels $L_\text{target}$. The targeted attack can be written as
\begin{equation}
    \Delta x = -\epsilon  \operatorname{sign}\left(\nabla_x \mathcal{L}(x,L_\text{target})\right).
\end{equation}

Both versions can be applied to TrOCR in a straightforward manner. As a loss function, we take the sum of the cross-entropy losses of the tokens, which was used in training. Then we can use backpropagation to compute the gradient with respect to the pixel values. 

\subsection{DeepFool}\label{Sec:Method/DeepFool}
The DeepFool algorithm \cite{moosavi2016deepfool} was developed as an extension to FGSM. It is an untargeted attack algorithm. The key idea is that, at each iteration, the decision boundaries are approximated by means of linearization. This creates a polyhedron. Then, the closest boundary is chosen and a perturbation to that boundary is made. This means that in each step we aim for the class to change to one other specific class, characterized by the smallest 
alteration. Iteration stops once the class changes. If we start with an image $x_0$ classified as $\hat{k}$, then for each iteration we first compute the following values for all classes:
\begin{align}
    w_k &\gets \nabla F_k(x_i) - \nabla F_{\hat{k}},\\
    f_k &\gets F_k (x_i) - F_{\hat{k}} (x_i).
\end{align}
Then the optimal class to target is 
\begin{equation}
    \overline{k} \gets \argmin_{k\neq \hat{k}} \frac{\vert f_k \vert}{\Vert w_k \Vert_2},
\end{equation}
and the resulting update is
\begin{equation}
    x_{i+1} \gets x_i + \frac{\vert f_{\overline{k}} \vert}{\Vert w_{\overline{k}} \Vert_2^2}w_{\overline{k}}.
\end{equation}    
In applying DeepFool to TrOCR, we need to consider decision boundaries for all tokens. We aim for all tokens to be mislabeled. Therefore, we 
could compute updates for every token consecutively. Because the number of possible tokens in the vocabulary is high, a maximum number of decision boundaries to check is chosen, using the classes with the highest logits. The overall method is given in \cref{Alg:DeepFool}. 

In practice, it was found that a single iteration is sufficient. Almost always, the closest decision boundary was that of the second most likely class. Therefore in testing we use topAmount $=1$ and maxIt $=1$ in \cref{Alg:DeepFool}.

\begin{algorithm}
\caption{DeepFool for TrOCR}\label{Alg:DeepFool}
\begin{algorithmic}[1]
    \State\textbf{Input:} $F,x,\text{topAmount, maxIt}$
    \State\textbf{Output:} $\Delta x$
    \State out $\gets F(x)$
    \State orgLabels, labels $\gets$ Labels(out)
    \State len $\gets$ Len(orgLabels)
    
    \State \textbf{initialize} $x_1 \gets x$ \textbf{ and } $i \gets 1$
    \While{$\text{Any}_j\left(\text{orgLabels}_j = \text{labels}_j\right)\textbf{and }  i \leq \text{maxIt}$}
        \For{$t \in \{1,\hdots,\text{len}\} \textbf{with } \text{orgLabels}_t = \text{labels}_t$}
            \State $\hat{k} \gets \text{orgLabels}_t$
            \State topLabels $\gets \operatorname{argTopK}_k\left(\text{out}^t_k, \text{topAmount}+1\right)$
            \For{$k\in \text{topLabels} \textbf{ with } k\neq \text{orgLabels}_t$}
                \State $w_k^t \gets \nabla F^t_k(x_i) - \nabla F^t_{\hat{k}}(x_i)$
                \State $f_k^t \gets F_k^t (x_i) - F^t_{\hat{k}} (x_i)$
            \EndFor
            \State$\overline{k}_t \gets \argmin_{k\neq \hat{k}} \frac{\vert f_k^t \vert}{\Vert w_k^t\Vert_2}$
            \State$r^t_i \gets \frac{\vert f_{\overline{k}} \vert}{\Vert w_{\overline{k}} \Vert_2^2}w_{\overline{k}}$
        \EndFor
        \State $u_i \gets \sum_t r^t_i $
        \State $x_{i+1} \gets x_i + u_i$
        \State out $\gets F(x_{i+1})$
        \State labels $\gets$ Labels(out)
        \State $i \gets i+1$
    \EndWhile
    \State $\Delta x = \sum_i u_i$
    \State \Return{$\Delta x$}
\end{algorithmic}
\end{algorithm} 

\subsection{Backward Error}\label{Sec:Method/BE}
Adversarial attacks based on backward error analysis were first introduced in \cite{beuzeville:hal-03296180} and extended in \cite{beerens2023adversarial}. 
The idea is that at each iteration, the neural network is linearized around the current image, and a linear least-squares constrained optimization problem is solved to find the smallest perturbation leading to the desired misclassification. This optimization problem may then be solved by a state of the art solver;
we use OSQP \cite{osqp}.

To extend this to TrOCR, we may change the misclassification constraint to hold for all tokens that do not align with the targets. Instead of comparing the target label with all the options in the vocabulary, we will only compare with the current label. This is done to limit the amount of backpropagation. Finally, we also remove the constraint keeping the pixel values in $[0,1]$, to be more closely comparable with the other algorithms. The method is given in \cref{Alg:BE}. 

\begin{algorithm}
\caption{Backward Error attack for TrOCR}\label{Alg:BE}
\begin{algorithmic}[1]
    \State\textbf{Input:} $F,x,\alpha,$ target, iterations
    \State\textbf{Output:} $\Delta x$
    \State out $\gets F(x)$
    \State labels $\gets$ Labels(out)
    
    \State $x_1 \gets x$
    \State $\Delta x \gets 0$

    \State diff $\gets \{ t: \text{target}_t \neq \text{labels}_t \}$

    \For{$i = 1$ \textbf{to} iterations}
    
        \State $\delta x \gets$ Variable
        \State $y_\text{orig} \gets$ Variable
        \State $y_\text{targ} \gets$ Variable
        \State objective $\gets \text{Objective}( \Vert \Delta x + \delta x\Vert_2)$
        \State constr $\gets \text{Constraint}(y_\text{orig} \leq y_\text{targ})$

        \For{$j = 1$ \textbf{to} Len(diff)}
            \State $t\gets \text{diff}_j$
            \State $\text{orig} \gets\text{labels}_t$
            \State $\text{targ} \gets \text{target}_t$
            \State constr.add$\left(y_\text{orig}^j = \text{out}_\text{orig}^t + \nabla_x F^t_\text{orig}(x_i)\cdot \delta x\right)$
            \State constr.add$\left(y_\text{targ}^j = \text{out}_\text{targ}^t + \nabla_x F^t_\text{targ}(x_i)\cdot \delta x\right)$
        \EndFor
        \State $\delta x,y_\text{orig},y_\text{targ} \gets \text{Solve(objective, constr)}$ 
        \State $\Delta x \gets \Delta x + \alpha \delta x$
        \State $x_{i+1} \gets x_i + \alpha \delta x$
        \State out $\gets F(x_{i+1})$
        \State labels $\gets$ Labels(out)
    \EndFor

    \State \Return{$\Delta x$}
\end{algorithmic}
\end{algorithm}

\subsection{Carlini and Wagner}\label{Sec:Method/CW}
In contrast to FGSM, where the training loss function is used,
the Carlini and Wagner (C\&W) attack \cite{carlini2017towards} is based on the idea of minimizing a specially created loss function. 
It can be used in targeted and untargeted situations.  
When targeting a label $\hat{k}$, the new loss function is defined as
\begin{equation}
    \mathcal{L}(\Delta x) = \Vert \Delta x \Vert_2^2 + c \ f\left( x+\Delta x , \hat{k}\right),
\end{equation}
where the first term on the RHS controls the $\ell_2$ norm of the perturbation and the second term encourages the desired classification. The constant $c>0$ is a parameter that balances the two requirements. The function $f$ is chosen so that $f(x+\Delta x,\hat{k})\leq 0$ if and only if $\argmax_k F_k(x+\Delta x) = \hat{k}$. In particular, it is chosen to be
\begin{equation}
    f\left( x', \hat{k}\right) = \left(\max_{k\neq \hat{k}}\left(F_k(x')\right) - F_{\hat{k}}(x')\right)^+,
\end{equation}
where $(\cdot)^+ = \max\{0,\cdot\}$. In that case $\hat{k}$ is the target class. In the untargeted setting, we take $\hat{k}$ as the predicted class and instead define
\begin{equation}
    f\left( x', \hat{k}\right) = \left(F_{\hat{k}}(x') - \max_{k\neq \hat{k}}\left(F_k(x')\right)\right)^+.
\end{equation}
Then we have $f(x+\Delta x,\hat{k})\leq 0$ if and only if $\argmax_k F_k(x+\Delta x) \neq \hat{k}$. 

Pixels are represented by a change of variables, keeping them in the allowed range $[0,1]$. We represent perturbed image $x+\Delta x$ with values $w$ such that
\begin{equation}
    x+\Delta x = \frac{1}{2}\left(\tanh(w) + 1\right).
\end{equation}
Now the optimization problem becomes
\begin{equation}
    \argmin_w\left( \begin{aligned}
        &\left\Vert \frac{1}{2}\left(\tanh(w) + 1\right) - x \right\Vert^2_2 \\
        &\quad + c \ f\left( \frac{1}{2}\left(\tanh(w) + 1\right), \hat{k} \right)
    \end{aligned}\right),
\end{equation}
with the choice of $f$ depending on whether the attack is targeted. 
For each iteration, the loss function gradient is calculated and a step is taken based on the Adam optimization scheme \cite{kingma2014adam}.
    
With a suitable adjustment to $f$, we can use this approach against TrOCR. Now we have multiple tokens, so in the targeted case the function becomes
\begin{equation}
    f\left( x', L\right) = \sum_t\left(\max_{k\neq L^t}\left(F^t_k(x')\right) - F^t_{L^t}(x')\right)^+,
\end{equation}
where $L$ are the output labels. In the untargeted case the function will become 
\begin{equation}
    f\left( x', L\right) = \sum_t\left( F^t_{L^t}(x') - \max_{k\neq L^t}\left(F^t_k(x')\right) \right)^+,
\end{equation}
where $L$ are the target labels.

To limit the number of iterations, a maximum is set. Additionally, when the value of the loss function has not improved for five steps, the optimization process is terminated, and the perturbation with the lowest loss so far will be used as the output. In practice, when the targeted version was used, it was found that the loss tends to increase again after reaching the desired target. Therefore, we terminate immediately after reaching the target.

\subsection{Implementation}\label{Sec:Method/Implementation}
Some comments on the implementation are in order. All testing is done using PyTorch \cite{PyTorch19}. The TrOCR model is accessed through the Transformers package by Hugging Face \cite{Wolf_Transformers_State-of-the-Art_Natural_2020}. The image preprocessor included with TrOCR does not support backpropagation. To overcome this, we wrote an equivalent function that allows backpropagation. Similarly, the function used for inference does not support backpropagation, so 
we performed the inference steps manually. 

%% file: sec/4_experiment.tex
\section{Experiments}\label{Sec:Experiments}
\subsection{Evaluation Metrics}\label{Sec:Experiments/Eval}
The success of attacks in targeted and untargeted settings may be measured in several ways. In an untargeted setting, we simply 
wish to quantify the effect on performance. A standard benchmark for TrOCR is the IAM handwriting database test set \cite{marti2002iam}. This testing set consists of 1861 images of text written by 128 different writers. In line with a previous study of TrOCR \cite{li2023trocr}, we will score performance using the Character Error Rate (CER), with a smaller CER indicating better model performance. The attacks aim to make the CER higher. This measure is based on the number of substitutions, deletions, and insertions needed to recover the correct answer. Here, we define the correct answer to be the true label given by the dataset. We will 
report this performance score in terms of the perturbation size.

In the targeted setting, we will look at the percentage of the attacked images that are classified with the target labels. 

\subsection{Experimental Setup}
In the untargeted setting, we consider FGSM, DeepFool, and C\&W attacks. The resulting attacks are then evaluated at the same relative scales; that is, for any attack $\Delta x$ on an image $x$, we evaluate
\begin{equation}\label{eq:pertEval}
    F\left( x + \epsilon  \frac{\Vert x\Vert_2}{\Vert \Delta x\Vert_2} \Delta x \right),
\end{equation}
with values $\epsilon \in \{n/10^4 : n= 0,1,\hdots,99\}$. Then we find the labels with the highest logit scores and decode the resulting sequence into words. These can then be compared with the original decoded labels using the CER. This will give us graphs of the CER on those perturbation sizes for every algorithm. 

FGSM does not require any parameters to be set. DeepFool is used with a single iteration and with a single class other than the original. C\&W uses a learning rate and weight decay for Adam of $0.002$ and $10^{-5}$ respectively. These values were experimentally fine-tuned from the default values. The value of $c$ is $0.05$, based on the recommendation in \cite{carlini2017towards} to choose a low $c$ that leads to misclassification. As initialization, we use a perturbation that changes every pixel by $\eta = 0.00002$ in a random direction. This is then clipped, if necessary, to stay within the range $[0,1]$. To accommodate the change of variables, we then change every pixel at $0$ or $1$ to $\eta$ and $1-\eta$ respectively. The recommendation 
in \cite{carlini2017towards} 
that these initial perturbations should be about the size of the eventual perturbation inspired this choice of $\eta$. A maximum number of $30$ iterations is used. This is done because in the untargeted case the algorithm often easily finds perturbations leading to the desired misclassification, and then spends more iterations optimizing the perturbation size. 

In the targeted case, we compare FGSM, Backward Error, and C\&W attacks. For every image, we create a perturbation in which we aim to change a random token to the token in the vocabulary with the tenth highest logit score. We regard this as a fair comparison because it changes tokens which offer the same typical difficulty.

Here we again evaluate the perturbations as in \cref{eq:pertEval}, but now for $\epsilon\in\{n/10^5 : n= 0,1,\hdots,99,100,110,\hdots,390\}$. We look at smaller perturbations than in the untargeted case, because generally the perturbations made by the algorithms did not become more successful when scaled to be larger. Having finer resolution allows us to more accurately find all successful attacks. More $\epsilon$ are added for a better comparison with the C\&W attack. We will view the perturbations created by this attack separately and evaluate them in the size in which they are outputted by the algorithm. This is because scaling these leads to unsuccessful attacks.

Again, we decode the labels from the outputs into words. However, this time we will not look at the average CER with respect to the original labels. Instead, we compute what percentage has a CER of 0 with respect to the target labels for some perturbation up to size $\epsilon$. That is, for every $n_0\in\{0,1,\hdots,99,100,110,\hdots,390\}$ we compute
\begin{equation}
\frac{1}{I}\sum_i \mathbf{1}\left(\exists_{n\leq n_0} : \operatorname{CER}_{i,n} =0\right),
\end{equation}
where $I$ is the total number of images, $\mathbf{1}$ indicates whether a statement is true and
\begin{equation}
    \operatorname{CER}_{i,n} = \operatorname{CER}\left( \text{pertText}_i(n/10^5), \text{targetText}_i \right),
\end{equation}
is the CER between target text of image $i$ and the decoded text output from image $i$ with a perturbation of $\epsilon = n/10^5$.

The targeted FGSM algorithm does not require parameters. The Backward Error attack, denoted BE, will use 5 iterations with a step size of $\alpha=0.5$. C\&W uses the same Adam learning rate and weight decay as in the untargeted case. To reach the desired classifications, it is now necessary to increase $c$ to 15 and $\eta$ to $0.002$. The maximum number of iterations is increased to 50. 
Although not reached in most cases, we found that this slightly higher upper bound was occasionally useful in allowing the iteration to reach the target.

\subsection{Results}
\begin{figure}
    \centering
    \includegraphics[width = \linewidth]{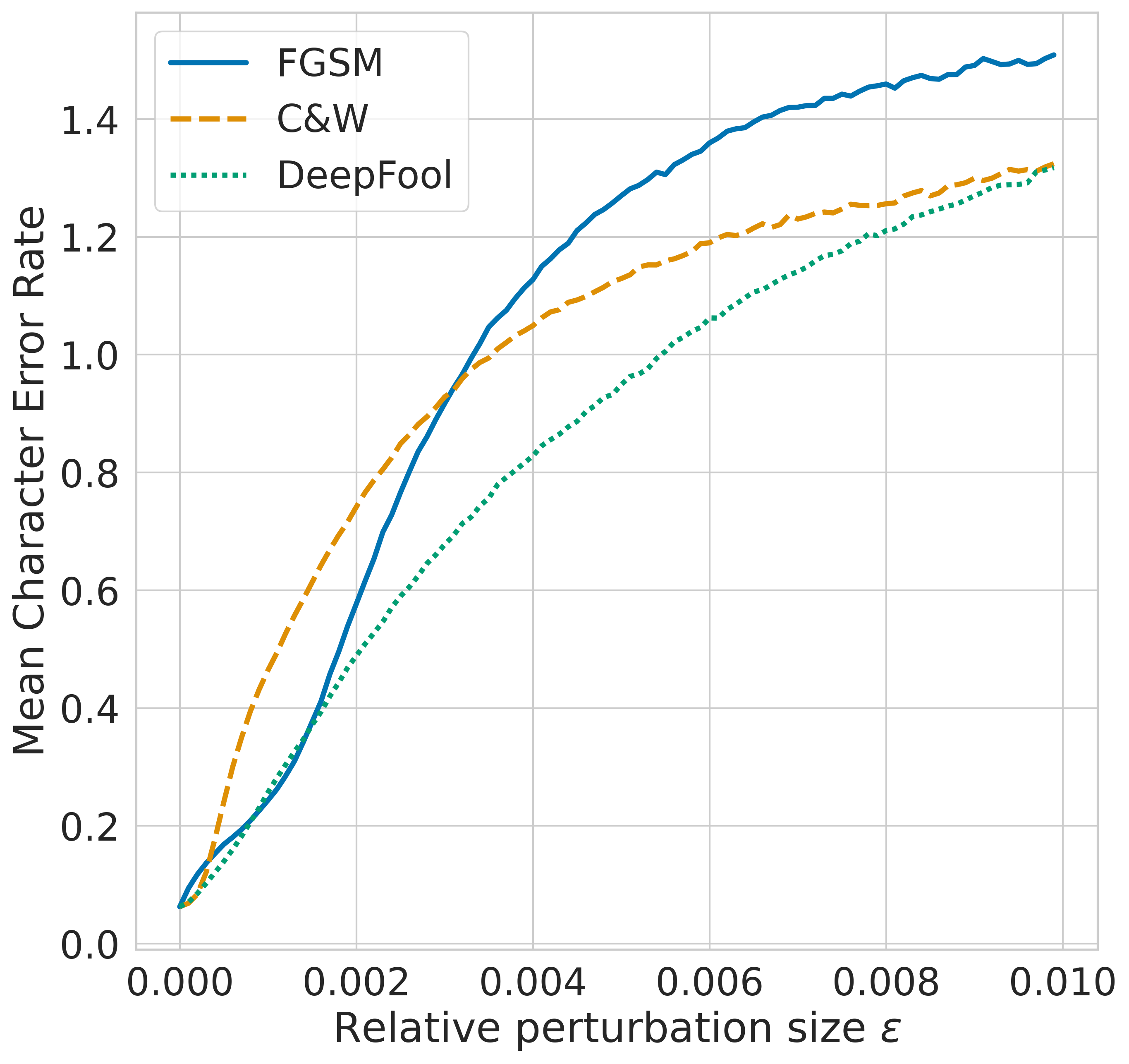}
    \caption{Comparison of mean Character Error Rate for adversarial attacks of different sizes generated by untargeted attacks FGSM, C\&W, and DeepFool on IAM handwriting database. }
    \label{fig:untargetedPlot}
\end{figure}

The mean CER values for the different untargeted attacks are shown in \cref{fig:untargetedPlot}. We notice that C\&W performs best for small $\epsilon$. This can be explained by the fact that it focuses on changing the token labels for the smallest perturbation. However, this token change might not change all characters in the words since two different tokens can contain overlapping characters. Also, FGSM is focusing on the cross-entropy loss, which looks at all the logits. Increasing this loss will initially not always lead to misclassifications of tokens. Further, it could be trying to increase the loss of one token more than that of the others. 

For larger perturbation sizes, FGSM becomes better than C\&W. This can be explained by C\&W being satisfied with any token change and optimizing the perturbation size for that change. Meanwhile, FGSM aims to increase the cross-entropy loss, which means that the logit for the actual class should decrease as much as possible. This will also lead to similar tokens not getting very high values. Also note that FGSM only overtakes when the CER is higher than 1. For such a high CER, insertions are made to make the sentences longer. 

In this context, 
DeepFool performs the worst of the three algorithms for perturbations of all sizes. For larger perturbations, it suffers from the attack only considering a token misclassification, just like C\&W. Since DeepFool tries to make the smallest step to the second-best class based on a linearization, it might be too small a step for multiple tokens when we consider small perturbations. It does not focus well on any of the two aspects that make the other two algorithms work well in their own respect. 

\begin{figure}
    \centering
    \includegraphics[width=\linewidth]{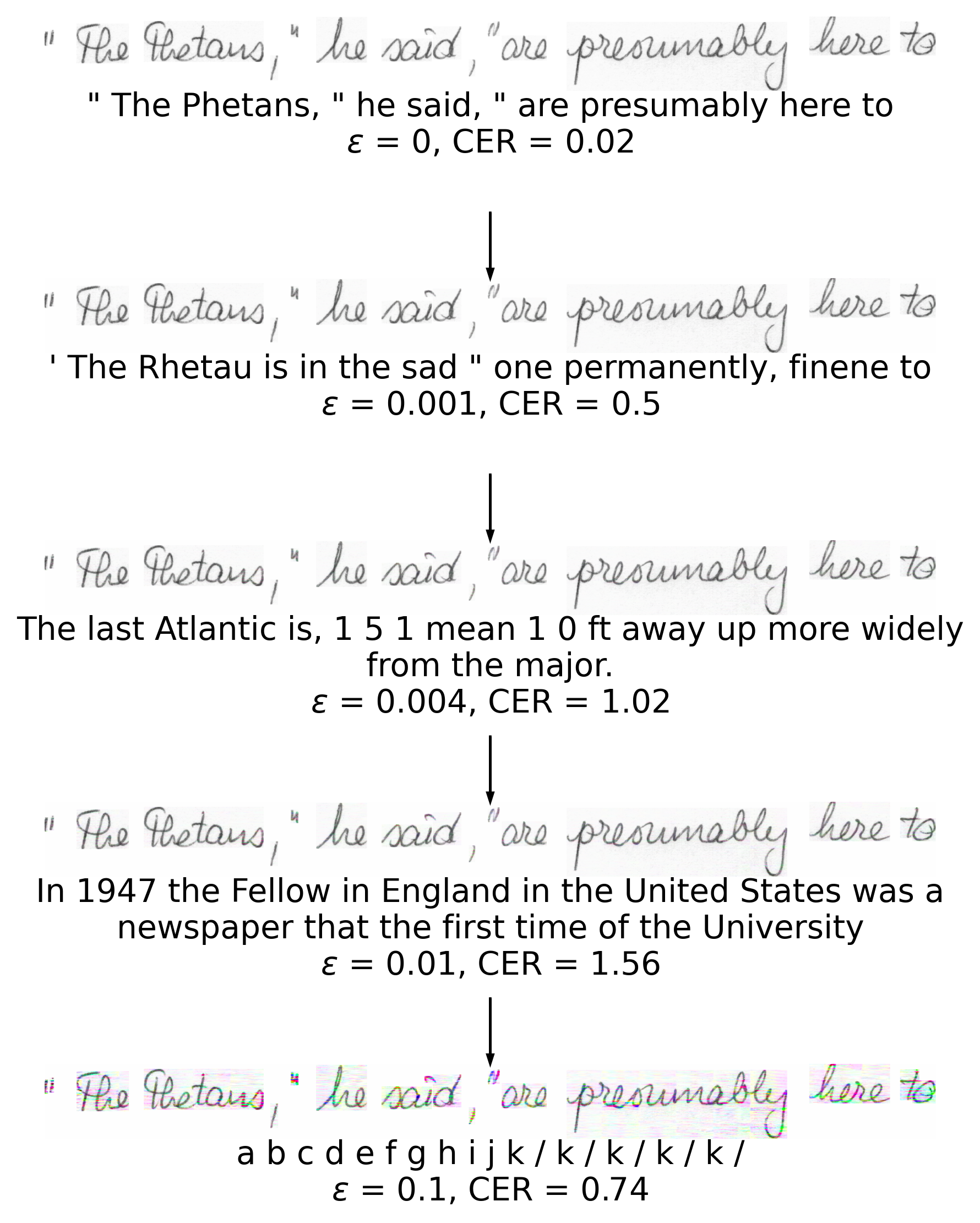}
    \caption{Perturbed images created by C\&W algorithm. From top to bottom, we start with the unperturbed image and gradually increase the relative norm of the perturbation. Each image evaluated with TrOCR. The output sentences are mentioned below the images. The correct text is '" The Thetans, " he said, " are presumably here to'.}
    \label{fig:compareCWPerts}
\end{figure}

To get a sense of the difference in scale of the perturbation sizes and the CER scores, we show an example of a C\&W attack in \cref{fig:compareCWPerts}. 
We show the same attack rescaled for different values of $\epsilon$. The correct sentence is `" The Thetans, " he said, " are presumably here to'. For $\epsilon = 0.001$, we see that the words in the output sentence look fairly similar to the words in the original image. For instance, `Thetans' becomes `Rhetau' and `are' becomes `one'. 
For the larger $\epsilon$ values in the figure, the output text no longer matches the original image. It is interesting that the output for $\epsilon=0.01$ consists of words that have some kind of mutual coherence. For $\epsilon=0.1$, TrOCR states part of the alphabet and gets caught in a loop where it outputs `k /' repeatedly. The CER generally increases as $\epsilon$ increases, but decreases for $\epsilon = 0.1$. This can be explained by the fact that shorter single-letter tokens need fewer insertions to create, which is the main driver of CER for large $\epsilon$ values. The image perturbations are 
essentially invisible for all but the largest $\epsilon$. 

\begin{figure}
    \centering
    \includegraphics[width=\linewidth]{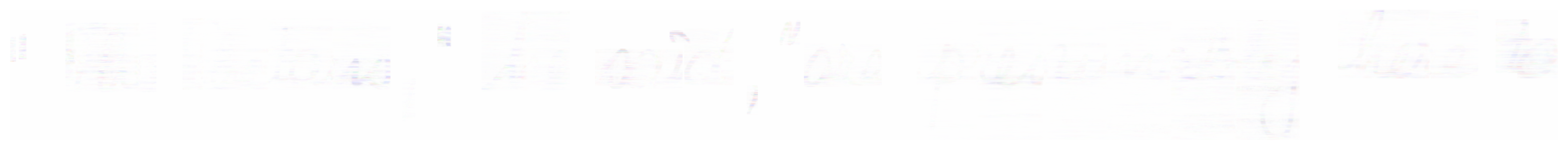}
    \caption{Almost imperceptible element-wise absolute value of the perturbation from \cref{fig:compareCWPerts} with $\epsilon=0.01$.}
    \label{fig:abspert}
\end{figure}

For further illustration, 
in \cref{fig:abspert}, we show the absolute values of the perturbation 
in \cref{fig:compareCWPerts} 
of size $\epsilon =0.01$ against a white background.
We see only a vague indication of the perturbation; it is almost imperceptible.
We also note that a CER of 1 was already reached with smaller perturbations.

From these tests, it can be concluded that TrOCR is highly vulnerable to untargeted attacks. The output can be completely changed to nonsensical text with
a perturbation that is not visible to the human eye.

\begin{figure}
    \centering
    \includegraphics[width=\linewidth]{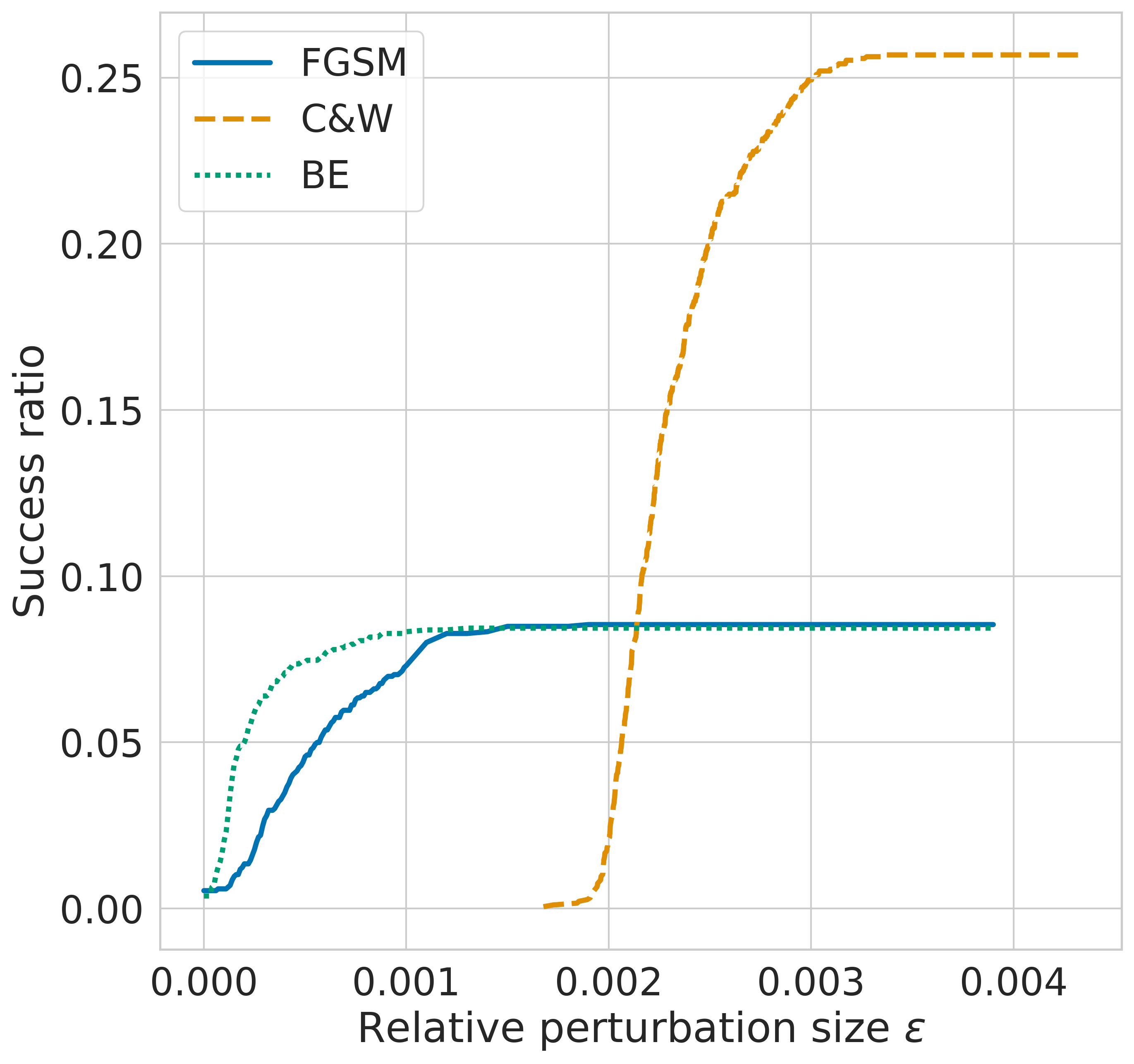}
    \caption{Success ratio for targeted attacks created by FGSM, C\&W, and BE on IAM handwriting database test set. The target is to change a random token position to the tenth most probable token from the vocabulary.}
    \label{fig:successRatios}
\end{figure}

For the targeted case, \cref{fig:successRatios} shows that C\&W needs significantly larger perturbation sizes to successfully reach the target labels, but it can successfully perturb $25.7\%$ of the images. This is significantly higher than FGSM and BE, which are able to successfully perturb only $8.5\%$ and $8.4\%$ respectively.
FGSM performs worse than BE, but quickly catches up as $\epsilon$ increases.

The fact that C\&W has larger perturbation sizes can be attributed to the initial perturbation. A larger starting perturbation is needed to increase the number of successful attacks, but the algorithm is not very good at decreasing the size of the perturbation. Successful attacks are done using perturbations of sizes close to $\epsilon=0.003$. In \cref{fig:compareCWPerts} it can be seen that this size of perturbation is not noticeable. Therefore, it is preferable to have higher success rates for the cost of the perturbations being this size.

In these tests, TrOCR is seen to be somewhat vulnerable to targeted attacks, but less so than it is to untargeted attacks. 
Note that in the targeted case, the size of the vocabulary complicates the choice of objective---we chose to compute results for the tenth most likely class as a representative example, but it would be of interest to consider other options.

%% file: sec/5_conclusion.tex
\section{Conclusion}
\label{Sec:Conclusion}
In this study, we 
devised the first range of attack strategies for the TrOCR model and 
conducted a comprehensive evaluation. 

In untargeted settings, we observed the vulnerability of TrOCR to adversarial attacks, particularly highlighting the efficacy of FGSM and C\&W attacks. C\&W demonstrated superior performance for small perturbation sizes, emphasizing its ability to focus on token changes. However, as perturbation sizes increased, FGSM outperformed C\&W, showcasing the nuances in attack strategies and their impact on TrOCR performance.

The results further revealed the susceptibility of TrOCR to targeted attacks, where we specifically targeted the tenth most likely token within the vocabulary. C\&W, although requiring larger perturbation sizes, achieved a success rate of approximately 25\%, outperforming the other algorithms. 

TrOCR uniquely combines the advantages of CV and NLP models to create a powerful OCR process.
However, our study demonstrates that 
TrOCR also inherits the vulnerabilities of the components that make up the overall computational pipeline.
This raises immediate security concerns, especially in high-risk applications such as finance, law and education,
that must be 
understood and addressed. 
Our results are also highly relevant for the current activity around regulation of AI.
Clearly, for any new regulations to be meaningful and realistic, they must be informed by results about current algorithmic limitations.

Finally, we note that since our attack strategies build on established methodologies, there is potential to 
adapt existing defense strategies \cite{yuan2019adversarial}, including adversarial training 
\cite{mmstv18}.
We therefore hope that this work motivates further research into the development of OCR systems that are both powerful and resilient.